\documentclass[10pt,twocolumn,letterpaper]{article}

\usepackage{iccv}
\usepackage{times}
\usepackage{epsfig}
\usepackage{graphicx}
\usepackage{amsmath}
\usepackage{amssymb}
\usepackage[square,sort,comma,numbers]{natbib}
\usepackage{graphicx}
\usepackage{amsmath}
\usepackage{amssymb}
\usepackage{float}
\usepackage{array}
\usepackage{pifont}
\usepackage{bbold}
\usepackage{subfigure}
\usepackage[ruled,vlined]{algorithm2e}
\usepackage{url}

\newcolumntype{P}[1]{>{\centering\arraybackslash}p{#1}}

\usepackage[pagebackref=true,breaklinks=true,letterpaper=true,colorlinks,bookmarks=false]{hyperref}

\iccvfinalcopy 

\def\iccvPaperID{1863} 
\def\httilde{\mbox{\tt\raisebox{-.5ex}{\symbol{126}}}}

\ificcvfinal\fi

\DeclareMathOperator*{\argmin}{arg\,min}

\DeclareMathOperator*{\maximize}{maximize}
\DeclareMathOperator*{\minimize}{minimize}

\begin{document}

\def\httilde{\mbox{\tt\raisebox{-.5ex}{\symbol{126}}}}
\def\balpha{\mbox{\boldmath $\alpha$}}
\def\bDelta{{\bf \Delta}}
\def\bLambda{{\bf \Lambda}}
\def\bvarphi{{\bf \varphi}}
\def\bTheta{{\bf \Theta}}
\def\btheta{\mbox{\boldmath $\theta$}}

\def\bPhi{\mbox{\boldmath{$\Phi$}}}
\def\vbphi{\vec{\mbox{\boldmath $\phi$}}}
\def\bb{{\bf b}}
\def\h{{\bf h}}
\def\bd{{\bf d}}
\def\ba{{\bf a}}
\def\bc{{\bf c}}
\def\p{{\bf p}}
\def\e{{\bf e}}
\def\s{{\bf s}}
\def\X{{\bf X}}
\def\x{{\bf x}}
\def\Y{{\bf Y}}
\def\y{{\bf y}}
\def\K{{\bf K}}
\def\k{{\bf k}}
\def\p{{\bf p}}
\def\bc{{\bf c}}
\def\A{{\bf A}}
\def\B{{\bf B}}
\def\C{{\bf C}}
\def\V{{\bf V}}
\def\S{{\bf S}}
\def\T{{\bf T}}
\def\W{{\bf W}}
\def\I{{\bf I}}
\def\U{{\bf U}}
\def\g{{\bf g}}
\def\G{{\bf G}}
\def\Q{{\bf Q}}
\def\d{{\bf d}}
\def\eg{{\it e.g.}}
\def\etal{{\it et. al}}
\def\H{{\bf H}}
\def\cR{{\bf R}}
\def\J{{\bf J}}
\def\bt{{\bf t}}
\def\bv{{\bf v}}
\def\R{{\bf R}}

\def\balpha{\mbox{\boldmath $\alpha$}}
\def\bdelta{\mbox{\boldmath $\delta$}}
\def\bzeta{\mbox{\boldmath $\zeta$}}
\def\bphi{\mbox{\boldmath $\phi$}}
\def\btau{\mbox{\boldmath $\tau$}}
\def\bmu{\mbox{\boldmath $\mu$}}
\def\bsigma{\mbox{\boldmath $\sigma$}}
\def\bSigma{{\bm \Sigma} }
\def\btheta{\mbox{\boldmath $\theta$}}
\def\dbphi{\dot{\mbox{\boldmath $\phi$}}}
\def\dbtau{\dot{\mbox{\boldmath $\tau$}}}
\def\dbtheta{\dot{\mbox{\boldmath $\theta$}}}
\def\bGamma{\mbox{\boldmath $\Gamma$}}
\def\bDelta{\mbox{\boldmath $\Delta$}}
\def\blambda{\mbox{\boldmath $\lambda $}}
\def\bOmega{\mbox{\boldmath $\Omega $}}
\def\bbeta{\mbox{\boldmath $\beta $}}
\def\bupsilon{\mbox{\boldmath $\Upsilon$}}
\def\myphi{\phi}
\def\bPhi{\mbox{\boldmath{$\Phi$}}}
\def\bLambda{\mbox{\boldmath{$\Lambda$}}}
\def\bSigma{\mbox{\boldmath{$\Sigma$}}}

\def\balpha{\mbox{\boldmath{$\alpha$}}}
\def\bbeta{\mbox{\boldmath{$\beta$}}}
\def\bdelta{\mbox{\boldmath{$\delta$}}}
\def\bgamma{\mbox{\boldmath{$\gamma$}}}
\def\blambda{\mbox{\boldmath{$\lambda$}}}
\def\bsigma{\mbox{\boldmath{$\sigma$}}}
\def\btheta{\mbox{\boldmath{$\theta$}}}
\def\bomega{\mbox{\boldmath{$\omega$}}}
\def\bxi{\mbox{\boldmath{$\xi$}}}

\def\bigO2{\mbox{${\cal O}$}}
\def\bigO{O}

\newcommand{\bH}{\mathbf{H}}
\def\mA{\mathcal{A}}
\def\mB{\mathcal{B}}
\def\mC{\mathcal{C}}
\def\mD{\mathcal{D}}
\def\mG{\mathcal{G}}
\def\mV{\mathcal{V}}
\def\mE{\mathcal{E}}
\def\mF{\mathcal{F}}
\def\mH{\mathcal{H}}
\def\mL{\mathcal{L}}
\def\mM{\mathcal{M}}
\def\mN{\mathcal{N}}
\def\mK{\mathcal{K}}
\def\mR{\mathcal{R}}
\def\mS{\mathcal{S}}
\def\mT{\mathcal{T}}
\def\mU{\mathcal{U}}
\def\mW{\mathcal{W}}
\def\mX{\mathcal{X}}
\def\mY{\mathcal{Y}}
\def\1n{\mathbf{1}_n}
\def\0{\mathbf{0}}
\def\1{\mathbf{1}}
\def\etal{{\em et al.}}

\def\balpha{\mbox{\boldmath $\alpha$}}
\def\bdelta{\mbox{\boldmath $\delta$}}
\def\bzeta{\mbox{\boldmath $\zeta$}}
\def\bphi{\mbox{\boldmath $\phi$}}
\def\btau{\mbox{\boldmath $\tau$}}
\def\bmu{\mbox{\boldmath $\mu$}}
\def\bsigma{\mbox{\boldmath $\sigma$}}
\def\bSigma{{\bm \Sigma} }
\def\btheta{\mbox{\boldmath $\theta$}}
\def\dbphi{\dot{\mbox{\boldmath $\phi$}}}
\def\dbtau{\dot{\mbox{\boldmath $\tau$}}}
\def\dbtheta{\dot{\mbox{\boldmath $\theta$}}}
\def\bGamma{\mbox{\boldmath $\Gamma$}}
\def\bDelta{\mbox{\boldmath $\Delta$}}
\def\blambda{\mbox{\boldmath $\lambda $}}
\def\bOmega{\mbox{\boldmath $\Omega $}}
\def\bbeta{\mbox{\boldmath $\beta $}}
\def\bupsilon{\mbox{\boldmath $\Upsilon$}}
\def\myphi{\phi}
\def\bPhi{\mbox{\boldmath{$\Phi$}}}
\def\bLambda{\mbox{\boldmath{$\Lambda$}}}
\def\bSigma{\mbox{\boldmath{$\Sigma$}}}

\def\balpha{\mbox{\boldmath{$\alpha$}}}
\def\bbeta{\mbox{\boldmath{$\beta$}}}
\def\bdelta{\mbox{\boldmath{$\delta$}}}
\def\bgamma{\mbox{\boldmath{$\gamma$}}}
\def\blambda{\mbox{\boldmath{$\lambda$}}}
\def\bsigma{\mbox{\boldmath{$\sigma$}}}
\def\btheta{\mbox{\boldmath{$\theta$}}}
\def\bomega{\mbox{\boldmath{$\omega$}}}
\def\bxi{\mbox{\boldmath{$\xi$}}}

\def\bPsi{\mbox{\boldmath $\Psi $}}
\def\bone{\mbox{\bf 1}}
\def\bzero{\mbox{\bf 0}}

\def\WB{{\bf WB}}

\def\A{{\bf A}}
\def\B{{\bf B}}
\def\C{{\bf C}}
\def\D{{\bf D}}
\def\E{{\bf E}}
\def\F{{\bf F}}
\def\G{{\bf G}}
\def\H{{\bf H}}
\def\I{{\bf I}}
\def\J{{\bf J}}
\def\K{{\bf K}}
\def\L{{\bf L}}
\def\M{{\bf M}}
\def\N{{\bf N}}
\def\O{{\bf O}}
\def\P{{\bf P}}
\def\Q{{\bf Q}}
\def\R{{\bf R}}
\def\S{{\bf S}}
\def\T{{\bf T}}
\def\U{{\bf U}}
\def\V{{\bf V}}
\def\W{{\bf W}}
\def\X{{\bf X}}
\def\Y{{\bf Y}}
\def\Z{{\bf Z}}

\def\b{{\bf b}}
\def\bc{{\bf c}}
\def\bd{{\bf d}}
\def\e{{\bf e}}
\def\f{{\bf f}}
\def\g{{\bf g}}
\def\h{{\bf h}}
\def\i{{\bf i}}
\def\j{{\bf j}}
\def\k{{\bf k}}
\def\l{{\bf l}}
\def\m{{\bf m}}
\def\n{{\bf n}}
\def\o{{\bf o}}
\def\p{{\bf p}}
\def\q{{\bf q}}
\def\br{{\bf r}}
\def\s{{\bf s}}
\def\t{{\bf t}}
\def\u{{\bf u}}
\def\v{{\bf v}}
\def\w{{\bf w}}
\def\bx{{\bf x}}
\def\y{{\bf y}}
\def\z{{\bf z}}

\def\vbphi{\vec{\mbox{\boldmath $\phi$}}}
\def\vbtau{\vec{\mbox{\boldmath $\tau$}}}
\def\vbtheta{\vec{\mbox{\boldmath $\theta$}}}
\def\vI{\vec{\bf I}}
\def\vR{\vec{\bf R}}
\def\vV{\vec{\bf V}}

\def\mvec{\vec{m}}
\def\fvec{\vec{f}}
\def\appfvec{\vec{f}_k}
\def\avec{\vec{a}}
\def\bvec{\vec{b}}
\def\evec{\vec{e}}
\def\uvec{\vec{u}}
\def\xvec{\vec{x}}
\def\wvec{\vec{w}}
\def\gradvec{\vec{\nabla}}

\def\aM{\mbox{\bf a}_M}
\def\aS{\mbox{\bf a}_S}
\def\aO{\mbox{\bf a}_O}
\def\aL{\mbox{\bf a}_L}
\def\aP{\mbox{\bf a}_P}
\def\ai{\mbox{\bf a}_i}
\def\aj{\mbox{\bf a}_j}
\def\an{\mbox{\bf a}_n}
\def\a1{\mbox{\bf a}_1}
\def\a2{\mbox{\bf a}_2}
\def\a3{\mbox{\bf a}_3}
\def\a4{\mbox{\bf a}_4}

\def\sx{\mbox{\scriptsize\bf x}}
\def\st{\mbox{\scriptsize\bf t}}
\def\ss{\mbox{\scriptsize\bf s}}
\def\cR{{\cal R}}
\def\calD{{\cal D}}
\def\calS{{\cal S}}

\def\sigmae{\sigma}
\def\sigmam{\sigma}

\def\balpha{\mbox{\boldmath{$\alpha$}}}
\def\bbeta{\mbox{\boldmath{$\beta$}}}
\def\bdelta{\mbox{\boldmath{$\delta$}}}
\def\bgamma{\mbox{\boldmath{$\gamma$}}}
\def\blambda{\mbox{\boldmath{$\lambda$}}}
\def\bsigma{\mbox{\boldmath{$\sigma$}}}
\def\btheta{\mbox{\boldmath{$\theta$}}}
\def\bomega{\mbox{\boldmath{$\omega$}}}
\def\bxi{\mbox{\boldmath{$\xi$}}}

\def\dx{{\delta \x}}
\def\dref{{\d_{ref}}}
\def\px{{\partial \x}}
\def\fxp{\f(\x, \p)}

\def\dfp{\mathbf{d}(\mathbf{f}(\x,\mathbf{p}))}
\def\dfpk{\mathbf{d}(\mathbf{f}(\x,\mathbf{p}^k))}
\def\Ep{E(\mathbf{\d, \p})}
\newcommand{\deltap}[1]{\Delta^{#1}}
\newcommand{\Jp}[1]{\J^{#1}}
\newcommand{\Hp}[1]{\H^{#1}}
\newcommand{\Hpnewton}[1]{\H^{#1}_{nt}}
\newcommand{\Psip}[1]{\Psi^{#1}}
\newcommand{\Phip}[1]{\Phi^{#1}}
\newcommand{\dPsip}[1]{\Psi_{#1}}
\newcommand{\dPhip}[1]{\Phi_{#1}}

\newcommand{\dn}{\d_{s}}
\newcommand{\dc}{\d}
\newcommand{\dnxt}[1]{\dn(\f(\x, #1))}
\newcommand{\dcur}[1]{\dc(\f(\x, #1))}

\newcommand{\one}{\mathbf{1}}
\newcommand{\zero}{\mathbf{0}}
\newcommand{\real}{\mathbb{R}}

\newcommand{\denselist}{\itemsep -1pt}
\newcommand{\sparselist}{\itemsep 1pt}


\title{Unsupervised Video Understanding by Reconciliation of Posture Similarities}
\author{Timo Milbich, Miguel Bautista, Ekaterina
Sutter, Bj\"orn Ommer\\
Heidelberg Collaboratory for Image Processing \\
IWR, Heidelberg University, Germany\\
{\tt\small \{timo.milbich, miguel.bautista, ekaterina.sutter, bj\"orn.ommer\}@iwr.uni-heidelberg.de}
}

\maketitle
\thispagestyle{empty}


\begin{abstract}
Understanding human activity and being able to explain it in detail surpasses mere action classification by far in both complexity and value. The challenge is thus to describe an activity on the basis of its most fundamental constituents, the individual postures and their distinctive transitions. Supervised learning of such a fine-grained representation based on elementary poses is very tedious and does not scale. Therefore, we propose a completely unsupervised deep learning procedure based solely on video sequences, which starts from scratch without requiring pre-trained networks, predefined body models, or keypoints. A combinatorial sequence matching algorithm proposes relations between frames from subsets of the training data, while a CNN is reconciling the transitivity conflicts of the different subsets to learn a single concerted pose embedding despite changes in appearance across sequences. Without any manual annotation, the model learns a structured representation of postures and their temporal development. The model not only enables retrieval of similar postures but also temporal super-resolution. Additionally, based on a recurrent formulation, next frames can be synthesized.
\end{abstract}

\vspace{-3mm}

\section{Introduction}
The ability to understand human actions is of cardinal importance for our interaction with another. Explaining the activity of another person by observing only individual postures and their temporal transitions in a sequence of video frames has been a long-standing challenge in Computer Vision. There are numerous applications in problems like activity indexing and search \cite{activityretrieval1,activityretrieval2,activityretrieval3}, action prediction \cite{videoclassification,forecasting}, behavior understanding and transfer \cite{behaviour1,behaviour2}, abnormality detection \cite{abnormality}, and action synthesis and video generation \cite{actionsynth1,generation1,antic_iccv_15,rubio_pr_15}. Our goal is to learn human activity on the finest accessible level, i.e., individual poses, by capturing characteristic postures and the distinctive transitions between them solely based on videos without requiring any manual supervision or pre-defined body models. The underlying deep learning approach is unsupervised and starts from scratch with only the video data and without requiring tedious user input or pre-trained networks. Our approach to activity understanding, summarized in Fig. \ref{fig:first_page}
(cf. Supplementary\footnote{Video material demonstrating the different applications can be found under  \url{https://hciweb.iwr.uni-heidelberg.de/compvis/research/tmilbich_iccv17}}),exhibits the following characteristics:

\begin{figure}[!t]
\includegraphics[width=0.48\textwidth]{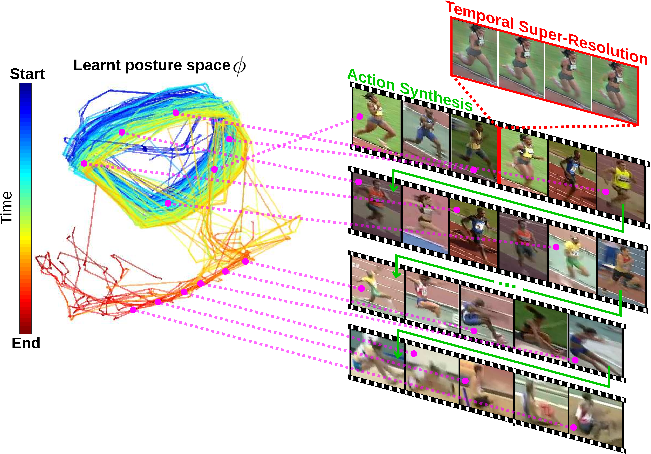}
\caption{Visualizing all frames of all long-jump sequences using the learnt posture representation $\phi$. Similar frames across different sequences and repetitions across time are mapped nearby, yielding a concise rendering of the overall activity with its characteristic gait cycles. Moreover, intermediate frames can be synthesized (top right) as well as future frames of a sequence (visualized by nearest neighbors from training set (right)).}
\label{fig:first_page}
\end{figure}

\textit{Unsupervised}: The most prominent paradigm to video understanding has been supervised action classification \cite{videoclassification}, since labeling finer entities such as individual poses \cite{mpii} is tedious. Action classification typically utilizes wholistic models and a discriminative approach is trained to classify actions into discrete classes. As a result, such approaches model actions globally in terms of their overall most salient differences, rather than capturing the subtle changes of human posture over time (the clothing of a person may suffcice to discriminate running from diving). 
\textit{Model-free}: Multiple works have modeled activity and posture using a predefined model for joint locations (i.e. MoCap \cite{mocap1,mocap2,haca}, Depth \cite{depth1}, etc.). However, obtaining this meta information is costly and prevents scaling these approaches to use large unlabeled collections of video data.
\textit{Continuous in time}: Several approaches have tackled the problem of understanding activity by decomposing it into discrete sub-actions \cite{mae1,actemes,mae2} or into a hierarchy \cite{haca,hierarchical1}. Consequently, the detailed, continuous evolution between consecutive postures is neglected. 
\textit{Multi-granular}: 
A lot of work has focussed separately either on pose matching \cite{PoseMatch,cliquecnn,deeppose}, action classification \cite{videoclassification,DenseTrajec,Action2Stream, antic_eccv_14}, or mid-level entities (e.g. clusters of postures) \cite{actemes,hierarchical1,mae1}. Explaining activities, however, demands to describe overall activity based on fine-granular postures and the transitions in between, thus linking coarse with fine granularities.
%
\textit{Fine-grained activity parsing}: Explaining activity on the temporal scale of single postures with all their diverse changes is far more detailed and complex than mere action classification \cite{DenseTrajec,Action2Stream}. Previous efforts \cite{storytelling} have approximated posture and its transitions using discrete states in an AND-OR graph and relied on tedious supervision information.

With no supervision information, no predefined model for human posture, and training from scratch, we need to compensate for there being no labels for individual postures. Since we are lacking the labels to directly train a representation for posture we can only utilize a large number of video frames and reason about pairwise relationships between postures. This is aggravated by the fact that the visual representation of posture in different videos can be significantly different due to changes in lighting, background, or the clothing and skin color of different persons. Therefore, we utilize a large number of training video frames and have a deep learning algorithm alternate between proposing pairwise similarities/dissimilarities between postures, and then resolving transitivity conflicts to bring the relationships into mutual agreement. To propose similarities, a combinatorial sequence matching algorithm is presented, which can find exact solutions, but only for small sets of frames. A CNN then resolves the transitivity conflicts between these different subsets of the training data by learning a posture embedding that reconciles the pairwise constraints from the different subsets. While the sequence matching is already incorporating information about posture changes, a Recurrent Neural Network (RNN) is trained to capture the overall activity and to predict future frames of an activity sequence by synthesizing transitions. 
\\
\indent
Experimental results show that our approach is able to successfully explain an activity by understanding how posture continuously changes over time and to model the temporal relationships between postures. Furthermore, our posture representation obtains state-of-the-art results on the problem of zero-shot human pose estimation and has also proven worth as a powerful initialization for other supervised human pose estimation methods. In addition, our approach captures the temporal progression of an activity, it can predict future frames, and it also enables a Generative Adversarial Network (GAN) to provide temporal super-resolution.

\section{Representation Learning for Parsing Activities}
\label{sec:learning_parsing}

\subsection{Learning a Posture Embedding}

We are interested in detailed understanding of human activity without requiring manual interaction or predefined models. Therefore, we can explain overall activity in a video only using the most basic entity that we can directly access: the human posture observed in bounding box detections $I$ of individual frames. Activity, which emerges at the temporal scale of an entire video sequence, is then represented by individual poses and their characteristic transitions and repetitions on a fine temporal scale. To model an activity, we need a posture representation $\phi(I; \theta)$ which: \emph{(i)} is invariant to changes in environmental conditions such as lighting and background. \emph{(ii)} is invariant to the appearance of persons (clothing and skin color). \emph{(iii)} is continuous in time (consecutive frames are near in feature space). Only then we can understand the essential characteristics of an activity and spot all repetitions of the same pose over time and in different sequences, despite changes in person appearance or environment. 
\\
\indent
A natural choice to incarnate $\phi(I; \theta)$ are CNNs. They exhibit great expressive power to learn highly non-linear representations at the cost of requiring millions of manually labeled samples for training. A popular alternative is then to only fine-tune a representation that has been pre-trained for discrete classification on large datasets \cite{imagenet}. Unfortunately, the performance of pre-trained models for transfer learning is heavily task dependent. In our scenario, the discrete classification objective contradicts our requirements for $\phi$, i.e., a discrete classification loss neglects smoothness within the representation space $\phi$. In addition, since datasets like \cite{imagenet} are composed of single images, pre-trained models fail to encode temporal relationships between frames. This explains the inferior performance of these pre-trained models in Sect. \ref{sect:experiments}.
\\
\indent
Instead of using a pre-trained representation we seek to learn a representation $\phi(I; \theta)$ that maps similar postures close in feature space while retaining temporal structure of an activity. Ideally, manual supervision of human postures, such as joint annotations, and/or positive links of similar postures within and across sequences together with negative links of dissimilar postures could be used to learn $\phi(I; \theta)$ employing, for example, triplets of similar and dissimilar poses. However, we are lacking these labels, altogether. To overcome this lack of labels we exploit the relationships inherent in large collections of video sequences. We infer the supervision information, which is required to learn a CNN representation $\phi(I; \theta)$, by solving a combinatorial sequence matching problem. The solution of this matching problem then provides us with correspondences of similar and dissimilar postures, which we then impose onto the CNN representation $\phi(I; \theta)$ to learn it.

\subsection{Sequence Matching for Self-supervision}

\begin{figure}[!t]
\includegraphics[width=0.45\textwidth]{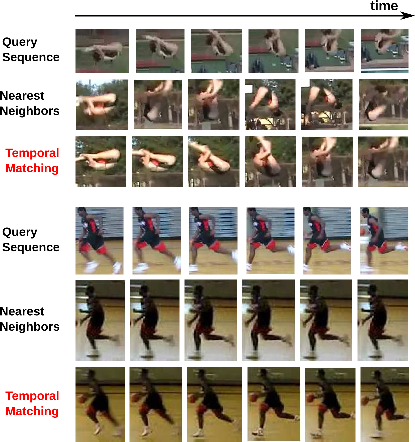}
\caption{Correspondences for two query sequences obtained using nearest neighbors (first row) and our sequence matching with temporal constraints (second row). Note how the temporal constraints provide much more accurate correspondences.}

\vspace{-3mm}

\label{fig:matching}

\end{figure}

We aim to learn a CNN representation $\phi(I; \theta)$ which encodes posture similarity, without being provided with any labels. In order to learn such a representation, we employ a self-supervision strategy, leveraging the temporal information in videos to solve a sequence matching problem and find pair-wise correspondences between frames on a sequence level. Let  $\mathcal{S} = \{I_j\}_{j=1}^{n}$ and $\mathcal{S}' = \{I_{j^{\prime}}^{\prime}\}_{j^{\prime}=1}^{n^{\prime}}$ denote two sequences of $n$ and $n'$ frames respectively, we want to find a correspondence $\pi:\{1,...,n\} \mapsto \{0,1,...,n^{\prime}\}$ that matches frames of $\mathcal{S}$ to frames of $\mathcal{S'}$, where the index $0$ is used to match outliers. Furthermore, in order to successfully learn $\phi(I; \theta)$ using self-supervision, we want to enforce the following constraints on $\pi$: $\emph{(i)}$ Corresponding frames should be similar in appearance. $\emph{(ii)}$ To avoid temporal cross-over and to reduce false positives matches, consecutive correspondences must be chronologically ordered. $\emph{(iii)}$ To avoid only part of a sequence being used and to explore the full span of possible postures therein, one-to-many correspondences should be penalized. $\emph{(iv)}$ Correspondences should be invariant to the sequence frame rate. Figure \ref{fig:matching} demonstrates the need for our temporal constraints. These constraints prevent, by definition, to utilize classical sequence matching approaches like the computational costly String Matching \cite{kmp} or Dynamic Time Warping \cite{dtw1,dtw2}. Thus, we define the following optimization problem which combines all these constraints,

\vspace{-4mm}

\begin{alignat}{2}
\label{eq:seqmatching}
    \begin{split}
    & \minimize_{\pi: \{1,...,n\} \mapsto \{0,1,...,n^{\prime}\}} \sum_{j=1}^{n} \left\Vert \phi(I_j; \theta) - \phi(I_{\pi(j)}'; \theta) \right\Vert_2^2 \\
    & + \lambda_1 \sum_{j=1}^{n-1} \mathbb{1}_{\pi(j) > \pi(j+1)} + \lambda_2 \sum_{j=1}^{n-1} \mathbb{1}_{\pi(j)=\pi(j+1)} \\
    & + \lambda_3 \sum_{j=1}^{n-1} \mathbb{1}_{\pi(j) + 1  < \pi(j+1)} \left[\pi(j+1) - \pi(j)\right] 
    \end{split}\; ,
\end{alignat}

where $\mathbb{1}(\cdot)$ denotes the indicator function and $\lambda_1, \lambda_2, \lambda_3$ penalize the violations of the different temporal constraints. The use of inequalities ensures constraint \emph{(iv)}. To solve the optimization problem in Eq. \eqref{eq:seqmatching} we convert it into an Integer Linear Program (ILP). In order to do so, we define a matrix $\Z \in \{0,1\}^{n \times n' \times n'}$, where $z_{j,j'_1,j'_2} := \mathbb{1}_{\pi(j) = j'_1 \wedge \pi(j+1) = j'_2} $. A non-zero $z$ indicates matches for two consecutive frames starting at position $j$. The ILP is then

\vspace{-4mm}

\begin{alignat}{2}
\centering
\label{eq:ilp}
\begin{split}
\maximize\limits_{ \Z \in \{0,1\}^{n \times n' \times n'} } & \sum_{j = 1}^{n-1} \sum_{j'_1,j'_2 = 0}^{n'}  z_{j,j'_1,j'_2} p_{j,j'_1,j'_2} \\
\text{subject to} & \sum_{j'_1,j'_2 = 0}^{n'} z_{j,j'_1,j'_2} = 1 \\
\wedge & \sum_{j'_1 = 0}^{n'} z_{j,j'_1,j'_2} = \sum_{j'_3 = 0}^{n'} z_{j+1,j'_2,j'_3}
\end{split}
\end{alignat}

where $p_{j,j'_1,j'_2}$ is the sum of all terms in Eq. \eqref{eq:seqmatching} with $z_{j,j'_1,j'_2} = 1$. To obtain reliable self-supervision information, we obtain an exact solution of this ILP using a branch-and-cut algorithm \cite{branchandcut}. However, exactly solving this problem for pairs of long sequences (e.g. $n>500$) is a costly operation with exponential worst case complexity in $n$, making it computationally infeasible.

To circumvent this high cost when matching $\mathcal{S}$ onto $\mathcal{S}'$, we break the target sequence $\mathcal{S}'$ into $k$ equal length sub-sequences of length $n' \approx 40$ and find an exact solution for Eq. \eqref{eq:ilp} on each local sub-sequence in parallel. Thus, the overall computational cost is reduced by a factor of $k$, making it a feasible to tackle long sequences. 
However we obtain only local correspondences at sub-sequence level and thus, discarding important relationships between different sub-sequences that compose the overall activity. To compensate for this shortcoming, we train a CNN with the different, local sub-sequence solutions in subsequent mini-batches as discussed in Sect. \ref{sec:joint}. The CNN then reconciles the local sub-sequence correspondences. Thus, we benefit from combining the computational feasibility of exact local sub-sequence matching with the power of stochastic CNN training, which aggregates lots of local observations in one concerted representation.

\subsection{From Local Correspondences to a Globally Consistent Posture Representation}
\label{sec:joint}

Our training procedure combines multiple exact solutions to local ILPs to obtain self-supervision for training a joint CNN representation $\phi(I;\theta)$ that reconciles all the sub-problems. Since our goal is to learn a CNN representation for encoding human activity in a fully unsupervised manner, the mini-batches for training are composed just by pairs of sequences $\{\mathcal{S}, \mathcal{S}'\}$.
We find these pairs by first randomly choosing $\mathcal{S}$ and sampling $\mathcal{S}'$ from a set of nearest neighbour sequences $\mathcal{S}_{\text{nn}}$ to $\mathcal{S}$, thus sorting out totally unrelated sequences. $\mathcal{S}_{\text{nn}}$ is constructed using simple sequence descriptors by temporally pooling similarities over all frames of a video.
After breaking $\mathcal{S}'$ into equal-length sub-sequences, the ILP in Eq. \eqref{eq:ilp} yields a solution $\Z^*$. These are exact pair-wise correspondences between $\mathcal{S}$ and a particular sub-sequence of $\mathcal{S}'$.  We then use these correspondences $\Z^*$ to generate triplets $\mathcal{T}_t$ using a triplet sampling $\{ \mathcal{S},\mathcal{S}',\Z^*\} \mapsto \{\mathcal{T}_t,\}_{t=1}^{T}$. Here $\mathcal{T}_t~=~\{I_a, I_+, I_-\}$ consists of a randomly sampled anchor image $I_a \in \mathcal{S}$ and its positive correspondence $I_+ = \pi(I_a) \in \mathcal{S}'$, together with a randomly chosen negative $I_- \in \mathcal{S}'$. We randomly sample negatives based on the $p$-th percentile of the similarity distribution of sequence $\mathcal{S}'$ to $I_+$. That is, we compute the similarity of $I_+$ to each frame of $\mathcal{S}'$, and sample negatives from frames with a lower similarity than the $p$-th percentile. We thus include hard negatives by decreasing $p$ over epochs. Note that by sampling positives and negatives from the same sequence and by comparing the same sequence with different other sequences, relationships within sequences are also implicitly established. Using this triplet self-supervision we update the CNN parameters $\theta$ via back-propagation and the triplet ranking loss \cite{ConvNetpretext2,ConvNetSimTriplet}.

\vspace{-4mm}

\begin{alignat}{2}
\label{eq:triplet_loss}
L(\{\mathcal{T}_t\}_{t=1}^{T}; \theta) =& \frac{1}{T}\sum\limits_{t=1}^{T} L'(\mathcal{T}_t; \theta)\\
\begin{split}
L'(\mathcal{T}_t; \theta) =&
\big[ \left\Vert \phi(I_a; \theta) - \phi(I_+; \theta) \right\Vert_2^2
\\
-& \left\Vert \phi(I_a; \theta) - \phi(I_-;\theta) \right\Vert_2^2 + \delta \big]_+
\end{split}
\end{alignat}

\begin{figure}[!t]
\includegraphics[width=0.45\textwidth]{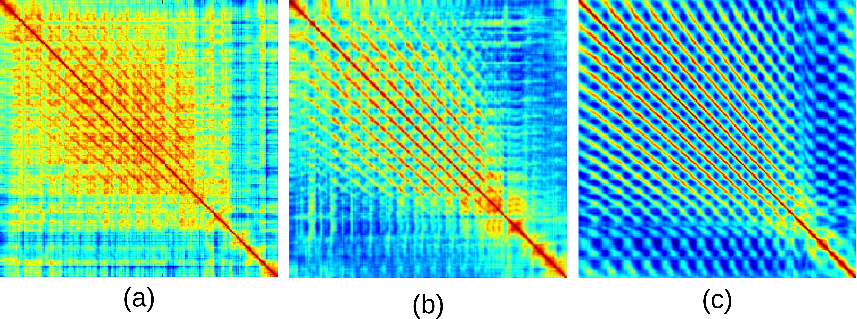}
\caption{Similarity matrices for a long jump sequence of Olympic Sports dataset computed using (a) VGG-S pre-trained on Imagenet, (b) CliqueCNN \cite{cliquecnn}, (c) Ours. The diagonal structures correspond to repetitions of gait cycles during running.}

\vspace{-3mm}

\label{fig:dist_mat}
\end{figure}

with $\delta$ controlling the margin between $I_+$ and $I_-$ with respect to $I_a$. The matching algorithm from Eq. \eqref{eq:ilp} provides the self-supervision information needed to train the CNN representation $\phi$, whereas the CNN training using Eq. \eqref{eq:triplet_loss} yields the posture embedding required to compute the similarities in Eq. \eqref{eq:ilp}. Alg. \ref{alg:training} outlines this iterative procedure. We found that using HOG-LDA \cite{hoglda} to initialize $\phi$ for the first epoch provides a decent initialization leading to a speed-up compared to a random $\phi$. Learning a single posture representation that captures characteristic similarities across and within sequences is thus decomposed into a series of mini-batch optimizations. For each, Eq. \eqref{eq:ilp} provides a matching that is locally, within the respective sub-sequences, optimal. The stochastic optimization of the CNN then consolidates, over a number of mini-batches, the transitivity conflicts between the local solutions to arrive at a single posture representation. That way, both approaches combine their strengths and weaknesses in an ideal manner. Fig.\ref{fig:dist_mat}(a-c) show an excerpt of similarity matrices from different models. Note the significantly improved signal-to-noise ratio in (c).

\subsection{RNN for Learning Temporal Transitions}
\label{sec:prediction}

\begin{algorithm}[!b]
\scriptsize
\KwData{$\{\mathcal{S}_i\}_{i=1}^{S}, \theta_{s=0}$} \tcp{Unlabeled video sequences and randomly initialized $\theta_{s=0}$}
\KwResult{$\{\phi,\theta\}$} 
\While{$\| \theta_{s+1} - \theta_{s} \|_2 > \epsilon$}{
    $( \mathcal{S}, \mathcal{S}') \leftarrow \{\mathcal{S}_i\}_{i=1}^{S}$ \tcp{Training batch}
    $\Z^* \leftarrow \argmin\limits_{\Z \in \{0,1\}^{n \times n' \times n'} }$ Eq. \eqref{eq:ilp} $(\mathcal{S},\mathcal{S}',\theta)$ 
    $\{\mathcal{T}_t\}_{t=1}^{T} \leftarrow \{\mathcal{S}, \mathcal{S}', \Z^*\}$ \tcp{Sample triplets}
    $\theta_{s+1} \leftarrow \theta_{s} + \alpha \nabla_{\theta_s} L(\{\mathcal{T}_t\}_{t=1}^{T}; \theta_s)$ \tcp{Update $\theta$}
} 
\caption{Unsupervised learning of a consistent posture representation using local correspondences.}
\label{alg:training}
\end{algorithm}

We now have an algorithm which effectively learns an overall posture representation $\phi(I;\theta)$ for an activity based on relationships within and across sequences. This detailed posture encoding is an ideal basis to also deal with coarser temporal scales, enabling to learn the complete temporal structure of an activity without requiring any prior model and to synthesize future frames of a sequence. We employ an RNN to explain or synthesize overall activity based on the transitions between individual postures. As we are living in a continuous state space with our representation $\phi$, also the transitions will be continuous. RNNs naturally lend themselves to encode temporally encoded entities, but significant modeling effort can be required to obtain a meaningful representation. Thus intensive pre-training of the static entities (such as words or objects) is used before learning transitions between them (sentences, videos) \cite{rnnword}. 

\begin{figure}[!t]
\includegraphics[width=0.45\textwidth]{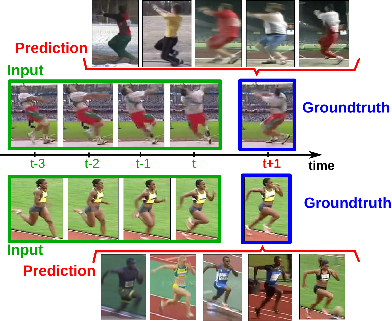}
\caption{Top $5$ LSTM predictions (red) for the next frame $t+1$, given four previous frames (green). Actual successor in blue.}
\label{fig:example_prediction}

\vspace{-5mm}

\end{figure}

However, the already structured posture space that we have learned in Sect. \ref{sec:joint} simplifies this and we can incarnate our RNN as a standard LSTM \cite{lstm}, which takes as input $\phi(I;\theta)$. In order for the LSTM to be able to model an activity as the temporal transition between individual postures, we formulate a regression task for future pose prediction under representation $\phi(I;\theta)$. By training the LSTM to predict the representation of future frames based on the observation of a small preceding sub-sequence, the network learns to model the evolution of the complete activity.
Let $\mathcal{C}^{t,l} = \{I_{t-l},\dots, I_{t}\} \subset \mathcal{S}$ be an $l$ frame sub-sequence of $\mathcal{S}$. The LSTM hidden state vector $\h_{t} \in \mathbb{R}^m$ has $m\gg d$ so as to effectively learn the variability of temporal transitions (cf. Sect. \ref{sect:experiments}). We then stack one fully connected layer on top of $\h$ to revert to the original dimensionality $d$, thus generating a representation $\phi'(\phi(I_t,\theta); \theta') \in \mathbb{R}^d$ (a stacking of an LSTM layer and a fully-connected layer). By formulating next frame prediction as a regression problem, we train this network using the euclidean loss function

\vspace{-4mm}

\begin{equation}
\label{eq:mse_loss}
L_{\text{RNN}}(\mathcal{C}^{t,l}, I_{t+1}, \phi, \theta; \theta') = \|\phi'(\phi(I_t,\theta); \theta')- \phi(I_{t+1}; \theta)\|_2^2
\end{equation}

The trained LSTM is then able to hypothesize transitions to future frames based on a small subsequence of an activity as demonstrated in Fig. \ref{fig:example_prediction}.

\section{Experimental Evaluation}\label{sect:experiments}

To evaluate our approach at all granularity levels of an activity, we report quantitative results for posture retrieval in Sect. \ref{sec:olympic_metrics} and human pose estimation (HPE) in Sects. \ref{sect:hpe}-\ref{sect:mpii}. In addition, we provide qualitative results for activity understanding in Sect. \ref{sect:tsne}, and for temporal super-resolution, and action synthesis in Sects. \ref{sect:temp_superres} to \ref{sect:synthesis}.

\subsection{Posture Retrieval: Olympic Sports dataset} \label{sec:olympic_metrics}

The Olympic Sports (OS) dataset \cite{olympic_sports} is a compilation of video sequences of $16$ different sports, containing more than $170000$ frames overall in $300$ video sequences. During training, each training mini-batch is generated by sampling two random sequences $\mathcal{S}, \mathcal{S}'$ and solving the ILP in Eq. \eqref{eq:ilp} to obtain correspondences $\Z^*$. Solving the ILP takes $\sim$ 0.1sec (IBM CPLEX Optimization framework) compared to $\sim$ 0.5sec to process a minibatch $\mathcal{T}_t$ on a NVIDIA Titan X (Pascal). We then use these correspondences to sample $\{\mathcal{T}_t, \}_{t=0}^{T=300}$ triplets, where the initial percentile $p=100$ is decreased by $10$ every epoch. In each frame the approach of \cite{dpm} yields person bounding boxes. During training we utilize VGG-S up to the \emph{fc6} layer and stack a $128$-dimensional \emph{fc7} layer on top together with an $l_2$-normalization layer, which is our representation $\phi(I;\theta)$. We use \emph{Caffe} \cite{caffe} for our implementation. 

\setlength{\tabcolsep}{1 pt}
\begin{table}
    \scriptsize
    \centering
    \begin{tabular}{|c|c|c|c|}
    \hline
    HOG-LDA \cite{hoglda}  & Ex-CNN \cite{exemplarcnn} &  VGG-S Imagenet \cite{vgg} & Doersch et. al \cite{ConvNetpretext1}    \\ \hline 
    0.62 &  0.56  &   0.64 & 0.58   \\ \hline \hline
    Shuffle\&Learn \cite{shuffleandlearn}  &  CliqueCNN \cite{cliquecnn} & Ours Scratch &  Ours Imagenet\\ \hline
    0.63 & 0.79 & \textbf{0.83} & \textbf{0.83} \\ \hline
    \end{tabular}
    \caption{Avg. AUC for each method on Olympic Sports dataset.}
    \label{tab:avg_auc}
    
    \vspace{-5mm}
\end{table}
\setlength{\tabcolsep}{5 pt}

To evaluate our representation on fine-grained posture retrieval we utilize the annotations provided by \cite{cliquecnn} and follow their evaluation protocol, using their annotations only for testing. We compare our method with CliqueCNN \cite{cliquecnn}, the triplet formulation of Shuffle\&Learn \cite{shuffleandlearn}, the tuple approach of Doersch et. al \cite{ConvNetpretext1}, VGG-S \cite{vgg}, and HOG-LDA \cite{hoglda}. For completeness we also include a version of our model that was initialized with Imagenet pre-trained weights \cite{vgg}. (\emph{i}) For CliqueCNN, Shuffle\& Learn, and Doersch et. al methods we use the models downloaded from their respective project websites. \emph{(ii)} Exemplar-CNN is trained using the best performing parameters reported in \cite{exemplarcnn} and the 64c5-128c5-256c5-512f architecture. Then we use the output of fc4 and compute 4-quadrant max pooling. During training of our approach, each image in the training set is augmented by performing random translation, scaling and rotation to improve invariance.
\\
\indent
In Tab. \ref{tab:avg_auc} we show the average AuC over all categories for the different methods. When compared with the best method so far \cite{cliquecnn}, the proposed approach improves the performance by $4\%$,  although the method in \cite{cliquecnn} was even pre-trained on Imagenet. This improvement is due to the cross sequence relationships enforced by our sequence matching, which enforce a representation which is invariant to background and environmental factors, encoding only posture. In addition, when compared to the state-of-the-art methods that leverage tuples \cite{ConvNetpretext1} or triplets \cite{shuffleandlearn} for training a CNN from scratch, our approach shows $20\%$ higher performance. This is explained by the more detailed similarity relationships encoded in the cross-sequence correspondences obtained by the sequence matching approach, which uses temporal constraints to obtain high quality relationships of similarity and dissimilarity. It is noteworthy, that our randomly initialized VGG-S trained with our self-supervision strategy yields equivalent performance to a version with pre-trained Imagenet weights as initialization. Thus, the proposed self-supervision circumvents the use of the $1.2$M labelled Imagenet samples. To the best of our knowledge, this is the first time that a self-supervised method performs equivalently without the widely adopted Imagenet pre-training strategy.

\subsection{Zero-Shot HPE}\label{sect:hpe}

After evaluating the proposed method for fine-grained posture retrieval, we tackle the problem of zero-shot pose estimation on the LSP dataset. That is, we transfer the pose representation learnt on Olympic Sports to the LSP dataset without any further training and retrieve similar poses based on their similarity. The LSP \cite{lsp} dataset is one of the most widely used  benchmarks for pose estimation. For evaluation we use the representation to compute visual similarities and find nearest neighbours to a query frame. Since the evaluation is zero-shot, joint labels are not available. At test time we therefore estimate the joint coordinates of a query person by finding the most similar frame from the training set and taking its joint coordinates. We then compare our method with VGG-S \cite{vgg} pre-trained on Imagenet, the triplet approach of Misra et. al (Shuffle\&Learn) \cite{shuffleandlearn} and CliqueCNN \cite{cliquecnn}. In addition, we also report an upper bound on the performance that can be achieved by zero-shot evaluation using ground-truth similarities. Here the most similar pose for a query is given by the frame which is closest in average distance of ground-truth pose annotations. This is the best one can achieve without a parametric model for pose (the performance gap to $100\%$ shows the discrepancy between poses in test and train set). For completeness, we compare with a fully supervised state-of-the-art approach for pose estimation \cite{chu17}. For computing similarities we now use the the intermediate \emph{pool5} layer of VGG-S as our representation $\phi(I;\theta)$, provided that our model is transferred from another dataset \cite{transferlearning}.
\\
\indent
In Tab. \ref{tab:results_lsp} we show the PCP@$0.5$ obtained by the different methods. For a fair comparison with CliqueCNN \cite{cliquecnn} (which was pre-trained on Imagenet), we include a version of our method trained using Imagenet initialization. Since in this experiment we are transferring our model from another dataset, we expect that Imagenet pre-training increases performance. Our approach significantly improves the visual similarities learned using both Imagenet pre-trained VGG-S and CliqueCNN \cite{cliquecnn}, obtaining a performance boost of at least $3\%$ in PCP score. In addition, when trained from scratch without any pre-training on Imagenet our model outperforms the  model of \cite{shuffleandlearn} by $13\%$, due to the fact that the cross-sequence correspondences obtained by our sequence matching approach encode finer relationships between samples. Finally, it is notable that even though our pose representation is \emph{transferred from a different dataset} without fine-tuning on LSP, it obtains state-of-the-art performance in the realm of unsupervised methods.

\begin{table}[!t]
    \scriptsize
    \centering
    \begin{tabular}{|c||c|c|c|c|c|c|c|}
    \hline
    Method  &T &UL &LL &UA &LA &H  &Total \\
    \hline \hline
    Ours Imagenet & 81.3 & 54.6 & 48.8 & 36.1 & 19.1 & 56.9 &  50.0\\
    \hline
    CliqueCNN \cite{cliquecnn}  & 80.1 & 50.1 & 45.7 & 27.2 & 12.6 &  45.5 &  43.5 \\
    \hline
    VGG-S \cite{vgg}& 82.0 & 48.2 & 41.8 & 32.4 & 15.8 & 53.6 & 47.0 \\
    \hline
    \hline
    Ours Scratch  & 73.0 & 45.1 & 41.6 & 26.2 & 12.2 & 44.4 & 43.0 \\
    \hline
    Shuffle\&Learn \cite{shuffleandlearn}  & 60.4  & 33.2 & 28.9 & 16.8 & 7.1 &  33.8 &  30.0\\
    \hline
    \hline
    Ground Truth & 93.7  & 78.8 & 74.9 & 58.7 & 36.4 & 72.4 & 69.2\\
    \hline
    \hline
    Chu et al. \cite{chu17} & 98.4 & 95.0 & 92.8 & 88.5 & 81.2 & 95.7 & 90.9 \\
    \hline
    \end{tabular}
    \caption{PCP measure for each method on Leeds Sports dataset for zero-shot pose estimation.}
    \label{tab:results_lsp}
    
    \vspace{-5mm}
\end{table}

\subsection{Self-supervision as Pre-training for HPE} \label{sect:mpii}

In addition to the zero-shot learning experiment we also evaluate our approach on the challenging MPII Pose dataset \cite{mpii} which is a state of the art benchmark for evaluation of articulated human pose estimation. The dataset includes around $25$K images containing over $40$K people with annotated body joints. MPII Pose is a particularly challenging dataset because of the clutter, occlusion and number of persons appearing in images. We are interested in evaluating how far our self-supervised approach can still boost a parametric approach that is trained with extensive supervision. Thus, we report the performance obtained by DeepPose \cite{deeppose}, when trained using as initialization each of the following models: Random initialization, Shuffle\&Learn \cite{shuffleandlearn}, Imagenet and our approach trained on OS (scratch and Imagenet pretraining). For this experiment the Alexnet \cite{alexnet} architecture is used like in \cite{deeppose}. Following the standard evaluation metric on MPII dataset, Tab. \ref{tab:results_mpii_deeppose} shows the PCKh@$0.5$ obtained by training DeepPose (stg-1) using their best reported parameters with the different initializations. 
\\
\indent
The performance obtained on MPII Pose benchmark shows that our unsupervised feature representation successfully scales to challenging datasets, successfully dealing with clutter, occlusions and multiple persons. In particular, when comparing our unsupervised initialization with a random initialization we obtain a $5.1\%$ performance boost, which indicates that our features encode a robust notion of pose that is robust to the clutter present in MPII dataset. Furthermore, we obtain a $1.2\% $ improvement over the  Shuffle\&Learn \cite{shuffleandlearn} approach. 

\begin{table}[!b]

    \vspace{-4mm}

    \scriptsize
    \centering
    \begin{tabular}{|p{13mm}||p{7mm}|p{5mm}|p{5mm}||p{8mm}|p{11mm}|}
    \hline
    & Ours scratch & S\&L \cite{shuffleandlearn} &  Rand. Init.  & Imagenet & Ours + Imagenet\\
    \hline \hline
    Head & 80.6 & 75.8 & 79.5 & 87.2 & 90.2\\
    \hline
    Neck  & 88.4 & 86.3 & 87.1 & 93.2 & 93.8\\
    \hline
    LR Shoulder & 74.8 & 75.0 & 71.6 & 85.2 & 86.3\\
    \hline
    LR Elbow. & 56.9 & 59.2 & 52.1 & 69.6 & 70.4\\
    \hline
    LR Wrist  & 41.6 & 42.2 & 34.6 & 52.0 & 58.6\\
    \hline
    LR Hip & 73.3 & 73.3 & 64.1 & 81.3 & 82.4\\
    \hline
    LR Knee & 63.6 & 63.1 & 58.3 & 69.7 & 73.2\\
    \hline
    LR Ankle & 56.9 & 51.7 & 51.2 & 62.0 & 67.4\\
    \hline
    Thorax & 88.6 & 87.1 & 85.5 & 93.4 & 93.7\\
    \hline
    Pelvis & 79.9 & 79.5 & 70.1 & 86.6 & 88.4\\
    \hline
    \hline
    Total & 70.5 & 69.3 & 65.4 & 78.0 & 80.4\\
    \hline
    \end{tabular}
    \caption{PCKh@$0.5$ measure on MPII Pose benchmark dataset using different initializations for the DeepPose approach \cite{deeppose}.}
    \label{tab:results_mpii_deeppose}
    
    \vspace{-5mm}
\end{table}

\subsection{Visualizing the Activity Representation}\label{sect:tsne}

\begin{figure}[!t]

\includegraphics[width=0.45\textwidth]{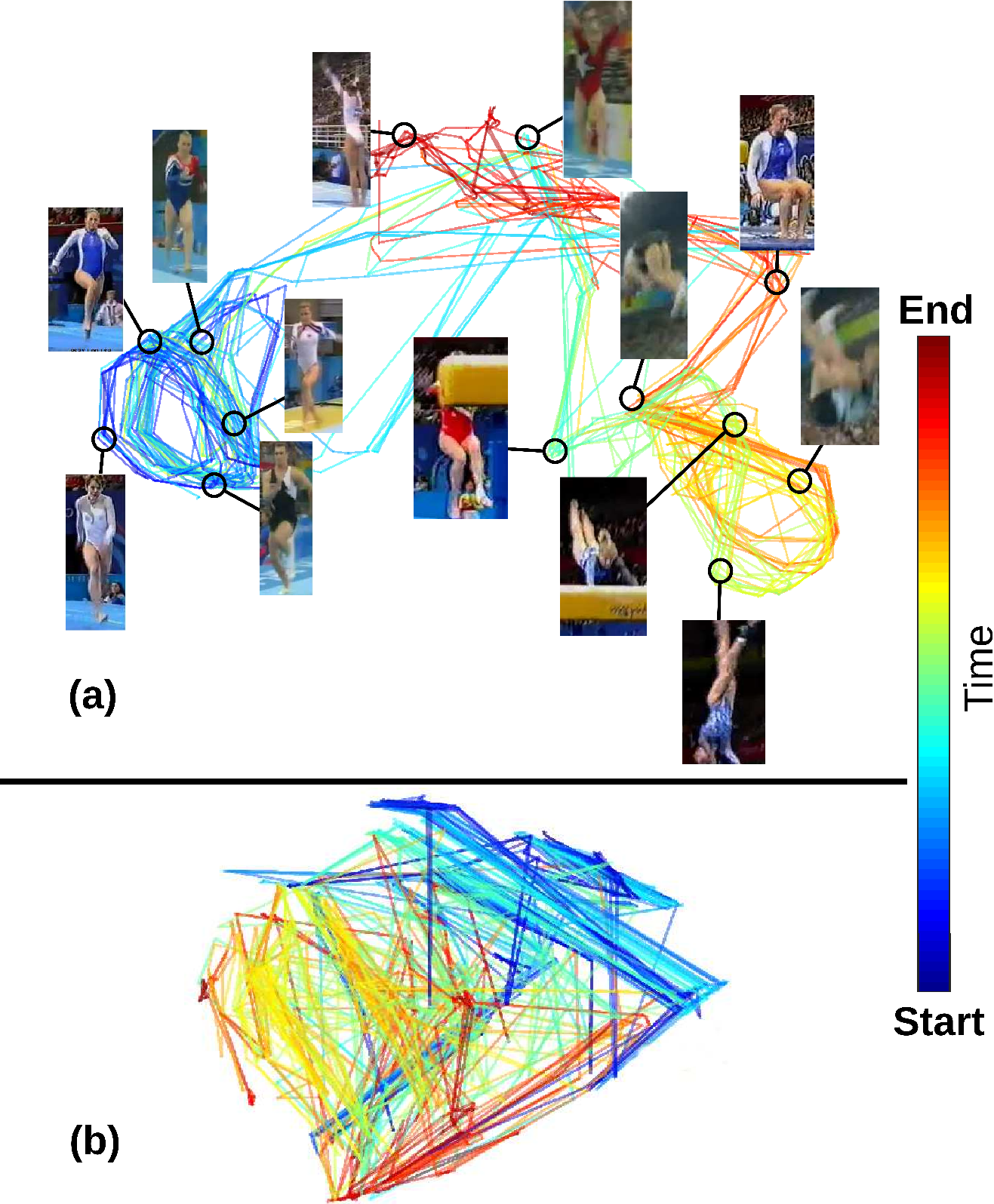}
\caption{Visualizing the learned posture representation $\phi$ and the progression of an activity (indicated by color). \emph{All} frames of \emph{all} vault sequences are shown (a) Our representation successfully learns the inherent structure of an action, e.g. repetitive gait and spinning cycles (blue, orange loops). Also, on a coarser temporal scale, repeated postures are brought near (outstretched arms before/after jump; cyan and red). Vector quantization of $\phi$ yields mutually dissimilar, characteristic poses shown in frames. (b) Representation obtained using \cite{shuffleandlearn}. As this model discards cross-sequence interactions, it misses the regularity of related postures across time and sequences.}
\label{fig:tsne_vault}

\vspace{-5mm}

\end{figure}

Whereas previous experiments have evaluated our representation on the level of individual poses, we now analyze the ability of $\phi(I;\theta)$ to also capture transitions between successive postures. Our model is trained using temporally aligned correspondences across sequences, thus representing not only relationships between different sequences but also encoding temporal transitions of pose within a sequence. Therefore, $\phi$ captures all the regularity of posture. It maps similar postures of different persons and repetitions of the same posture in a sequence, e.g., repeated gait cycles, to the same spot in feature space. Moreover, successive poses are also mapped to similar representations, so an activity has a smooth trajectory over $\phi$. To visually demonstrate the ability of $\phi(I;\theta)$ to capture the fine-grained pose interactions over time and between sequences we project the high dimensional representation $\phi$ to a 2D plot using the t-SNE procedure \cite{tsne}. Fig. \ref{fig:tsne_vault}(a) shows a mapping of all instances of vault activity. Successive postures within each video are connected by straight lines and color encodes the time within the sequence. The learned representation captures the repetitive structure of running and spinning (blue and orange loops) and the characteristic transitions between.
Additionally, the regularity of $\phi$ allows to group repetitive postures and to provide a condensed overview of an activity. Therefore, we employ standard agglomerative clustering \cite{agglomerative} to extract prominent mutually dissimilar posture that span an activity. We show the representative of each cluster on its corresponding location in Fig. \ref{fig:tsne_vault}(a). Moreover, we compare our representation with the state-of-the-art approach of Misra et al. \cite{shuffleandlearn} which introduces a temporal verification problem and learns to find the correct temporal order of triplets of postures within a sequence, Fig. \ref{fig:tsne_vault}(b). This shows that modeling only posture interactions within a sequence \cite{shuffleandlearn} and excluding cross-sequences correspondences as proposed in Sect. \ref{sec:learning_parsing} fails to capture the temporal evolution of an activity and degrades temporal structure.

\subsection{Inferring Temporal Super-Resolution}\label{sect:temp_superres}

\begin{figure}[!b]

\vspace{-5mm}

\includegraphics[width=0.48\textwidth]{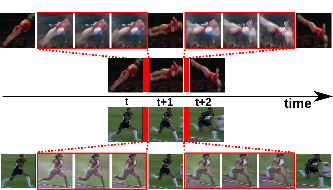}
\caption{Inferring intermediate frames between consecutive postures. For each transition we interpolate representations $\phi$ at regular intervals in between and invert interpolated features with \cite{deePSim}.}
\label{fig:super_res}

\vspace{-3mm}

\end{figure}

The previous section has demonstrated that our representation $\phi(I;\theta)$ successfully encodes posture and provides a basis for modeling the characteristic progression of an activity on the finest accessible level---individual frames. To further demonstrate the fine granularity at which activity is captured, we now unfold transitions between postures in consecutive frames, that is, we obtain super-resolution between two consecutive frames. Whereas \cite{irani} aggregates frames from within the same sequence, we bring all the different sequences with all their variability together. Since our representation maps related poses close to another, we can employ a local linear interpolation between successive postures to infer intermediate frames. Then the Generative Adversarial Network (GAN) of \cite{deePSim} is used to invert the feature back to an image. We use the implementation of \cite{deePSim} for both the generator and discriminator networks, where our learned activity representation $\phi(I;\theta)$ acts as the encoder network. To jointly train the three networks we use the DeePSiM-loss \cite{deePSim} considering adversarial and euclidean terms on both the image and feature domain. This inversion of our representation $\phi(I;\theta)$ creates images for the synthesized intermediate frames, allowing us to go past the limited temporal scale of given video sequences.  Fig. \ref{fig:super_res} shows temporal super-resolution results for two different activities. The continous progression of activity is preserved due to the continuity of our pose representation $\phi(I;\theta)$. It has finer temporal granularity than an individual video, since it interleaves a large number of related sequences, providing a truly continuous activity representation.

\subsection{Activity Understanding using LSTMs}

So far we have provided a comprehensive analysis of our activity representation, demonstrating its ability to understand actions on the fine-grained scale of single postures and beyond. Now, we evaluate the capability of our method to understand activity at the sequence level. Therefore, we train a recurrent network on top of the posture representation $\phi$ to yield a sequence-level encoding $\phi'$, as discussed in Sect. \ref{sec:prediction}. We employ an LSTM, trained on sub-sequences $\mathcal{C}^{t,l}$ of length $l=4$, sampled densely from all video sequences to predict the next succeeding frame $I_{t+1}$. During training, we sample mini-batches to cover the overall diversity of activity, so that all constituent postures are equally represented for learning the LSTM. Fig. \ref{fig:example_prediction} shows exemplary predictions from different activities.

\begin{figure}[!b]

\vspace{-4mm}

\centering
\begin{minipage}[c]{0.22\textwidth}
  \centering
  \includegraphics[width=\textwidth]{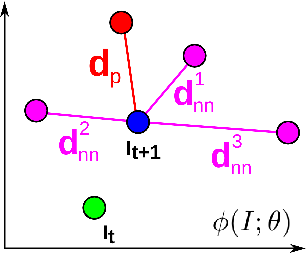}\\
  (a)
\end{minipage}
\begin{minipage}[c]{0.245\textwidth}
  \centering
  \includegraphics[width=\textwidth]{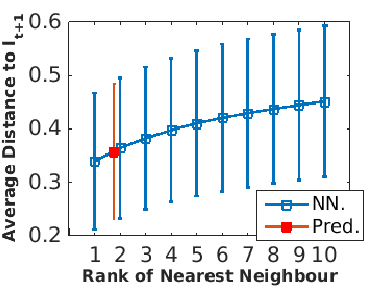}\\
  (b)
\end{minipage}
\caption{(a) Setup of quantitative evaluation. Blue is the actual next frame, red the prediction, purple the nearest neighbours. (b) LSTM evaluation: Comparing our average prediction error (red) against the average distance of $I_{t+1}$ to each of its $10$ nearest neighbours (blue). The error bars indicate the standard deviation of the measurements.} \label{fig:eval_prediction}
\end{figure}


Let us quantify the quality of predicted next frames. Given the true successor frame $I_{t+1}$, we identify its nearest neighbor in all the videos. We then compute the distance between these two frames and average it over all videos. The same is then done for the second, third, etc. nearest neighbor. Similarly, we compute the distance of our prediction and the true $I_{t+1}$ and also average it. Fig. \ref{fig:eval_prediction} (b) compares the resulting error of our prediction against that of the k nearearest neighbor from the dataset. Our prediction is better than actually observing the next frame and picking its second nearest neighbor. Despite the large variability of an activity this shows that the temporal progression of an activity has been well captured to yield favorable predictions of a successive frame.

\subsection{Video Understanding by Action synthesis} \label{sect:synthesis}

\begin{figure}[!t]
\centering
\includegraphics[width=0.45\textwidth]{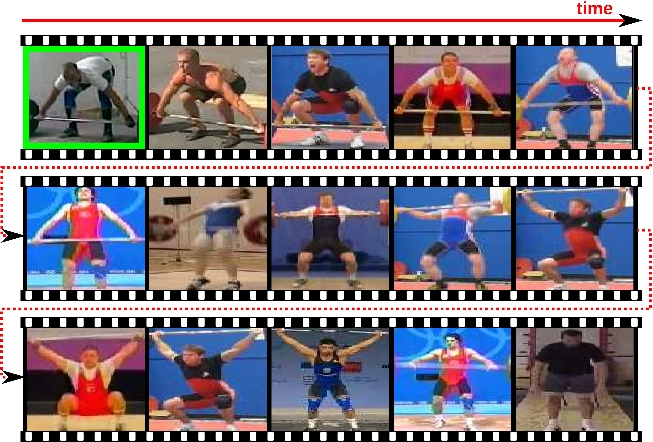}
\caption{Synthesizing an activity by recursively predicting the next posture. Green: initial image . Every $5$th frame is shown.}
\label{fig:synthesis}

\vspace{-5mm}

\end{figure}

We have just seen predictions of the next frame $I_{t+1}$ of a sequence. By recursively adding this predicted frame and then predicting a next successive frame we can iteratively synthesize an overall activity frame by frame. For visualization of the predicted next posture, we choose the nearest neighbor from the training set. Fig. \ref{fig:synthesis} summarizes the synthesis of a snatch activity initialized at the green posture. One can see that our model successfully infers the temporal ordering of the activity from its beginning until the end.

%

\section{Conclusions}

In this paper we have presented an unsupervised approach for understanding activity by means of its most fine-grained temporal constituents, individual human postures. A combinatorial sequence matching algorithm proposes relations between frames from subsets of the training set, which a CNN uses to learn a single concerted pose embedding that reconciles transitivity conflicts. Without any manual annotation, the model learns a structured representation of postures and their temporal development. The model not only enables retrieval of similar postures but also temporal super-resolution. Additionally, based on a recurrent formulation, future frames and activities can be synthesized.
\newline
\\
\noindent
\textbf{Acknowledgements:} This work has been supported in part by the Heidelberg Academy for the Sciences. We are grateful to the NVIDIA corporation for donating a Titan X GPU.

\bibliographystyle{plain}
\bibliography{egbib}
\end{document}